\documentclass[10pt,twocolumn,letterpaper]{article}

\usepackage{cvpr}
\usepackage{times}
\usepackage{epsfig}
\usepackage{graphicx}
\usepackage{amsmath}
\usepackage{amssymb}
\usepackage{subfigure}
\usepackage{url}

\usepackage[pagebackref=true,breaklinks=true,letterpaper=true,colorlinks,bookmarks=false]{hyperref}

\def\MH#1{{\color{red}{\bf [Mehrtash:} {\it{#1}}{\bf ]}}}

\def\GRASS#1#2{\mathcal{G}({#2},{#1})}

 \cvprfinalcopy 

\def\cvprPaperID{1192} 

\newcommand{\reals}[1]{\mathbb{R}^{#1}}
\newcommand{\enorm}[1]{\left\|{#1}\right\|}

\newcommand{\set}[1]{\left\{#1\right\}}

\DeclareMathOperator*{\trace}{Tr}
\newcommand{\half}{\frac{1}{2}}

\newcommand{\stiefel}{\mathcal{S}}
\DeclareMathOperator*{\subjectto}{\text{subject to}}
\newcommand{\eye}[1]{\mathbf{I}_{#1}}

\renewcommand\cdots{...}

\newcommand{\suptensor}[1]{\mathfrak{S}^{d}}

\newcommand{\fnorm}[1]{\left\|{#1}\right\|_F}

\newcommand{\comment}[1]{}
\newcommand{\ortho}{\mathcal{O}}
\newcommand{\grass}[1]{\mathcal{G}{#1}}

\DeclareMathOperator*{\minimize}{minimize}

\newcommand{\done}[1]{}
\newcommand{\seq}[1]{\langle {#1}\rangle}
\newcommand{\actodo}[1]{}
\newcommand{\bigoh}{O}
\ifcvprfinal\fi
\begin{document}

\title{Generalized Rank Pooling for Activity Recognition}

\author{Anoop Cherian$^{1,3}$\qquad Basura Fernando$^{1,3}$\qquad Mehrtash Harandi$^{2,3}$\qquad Stephen Gould$^{1,3}$\\
$^1$Australian Centre for Robotic Vision, $^2$Data61/CSIRO\\
$^3$The Australian National University, Canberra, Australia\\
{\tt\small firstname.lastname@\{anu.edu.au, data61.csiro.au\}}
}

\maketitle

\begin{abstract}
Most popular deep models for action recognition split video sequences into short sub-sequences consisting of a few frames; frame-based features are then pooled for recognizing the activity. Usually, this pooling step discards the temporal order of the frames, which could otherwise be used for better recognition. Towards this end, we propose a novel pooling method,~\emph{generalized rank pooling} (GRP), that takes as input, features from the intermediate layers of a CNN that is trained on tiny sub-sequences, and produces as output the parameters of a subspace  which (i) provides a low-rank approximation to the features and (ii) preserves their temporal order. We propose to use these parameters as a compact representation for the video sequence, which is then used in a classification setup. We formulate an objective for computing this subspace as a Riemannian optimization problem on the Grassmann manifold, and propose an efficient conjugate gradient scheme for solving it. Experiments on several activity recognition datasets show that our scheme leads to state-of-the-art performance. 
\end{abstract}
\section{Introduction}
\label{sec:intro}
Activity recognition from videos is challenging as real-world actions are often complex, confounded with background activities
and vary significantly from one actor to another. Efficient solutions to this difficult problem can facilitate several useful applications such as human-robot co-operation, visual surveillance, augmented reality, and medical monitoring systems. The recent resurgence of deep learning algorithms has demonstrated significant advancements in several fundamental problems in computer vision, including activity recognition. However, such solutions are still far from being practically useful and thus activity recognition continues to be a challenging research topic~\cite{aggarwal2011human,feichtenhofer2016convolutional,poppe2010survey,simonyan2014two,wang2013action,wang2015action}.

\begin{figure}
	\centering
	\includegraphics[width=8cm,trim={0cm 0cm 6cm 0cm},clip]{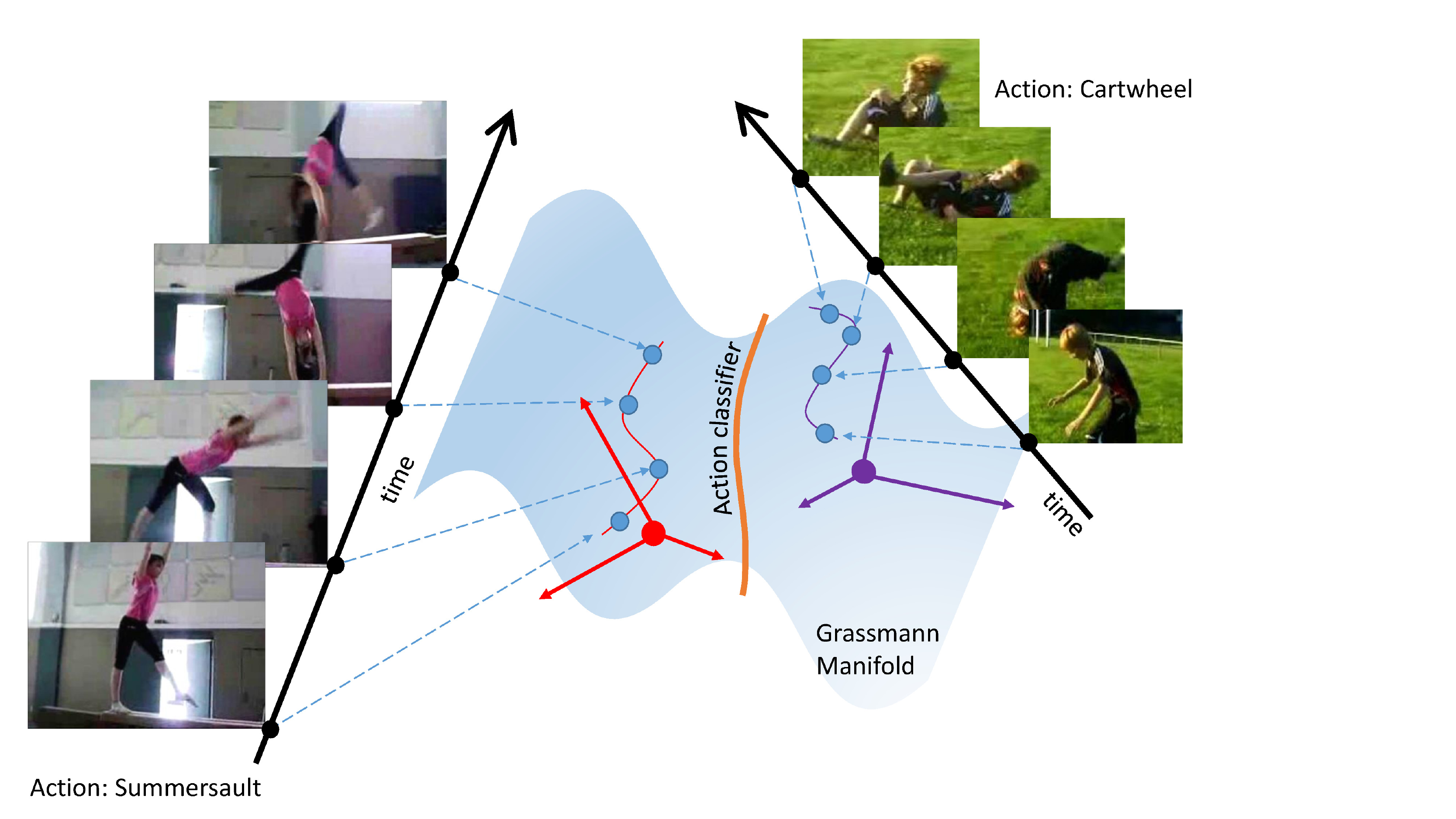}
	\label{fig:GRP_concept}
	\caption{An illustration of our pooling scheme. For every video, our formulation learns the parameters of a low-dimensional subspace in which the projected video frames conform to their temporal order. We use the subspaces as respective representations of the sequences. Such subspaces belong to the Grassmann manifold, on which we learn non-linear action classifiers.}
\end{figure}

Deep learning algorithms on long video sequences demand huge computational resources, such as GPU, memory, etc. One popular approach to circumvent this practical challenge is to train networks on  sub-sequences consisting of one to a few tens of video frames. The activity predictions from such short temporal receptive fields are then aggregated via a pooling step~\cite{karpathy2014large,donahue2014long,simonyan2014two}, such as computing the average or maximum of the generated CNN features. Given that the features are from temporally-ordered input data, it is likely that they capture the temporal evolution of the actions in the sequence. Thus, a pooling scheme that can use this temporal structure is preferred for activity recognition.

In Fernando et al.~\cite{fernando2015modeling,fernando2016discriminative}, the problem of pooling using the temporal structure is cast in a learning-to-rank setup, termed~\emph{Rank Pooling}, that computes a line in input space; the projection of input data onto this line preserving the temporal order. The parameters of this line are then used as a summarization of the video sequence. However,  several issues remain unanswered, namely (i) while the line is assumed to belong to the input space, there is no guarantee that it captures other properties of the data (other than order), such as background, context, etc. which may be useful for recognition, (ii) the ranking constraints are linear, (iii) each data channel (such as RGB) is assumed independent, and (iv) a single line for ordering is considered, while using multiple hyperplanes might lead to better characterization of the temporal action dynamics. In this paper, we propose a novel re-formulation of rank pooling that addresses all these drawbacks.

Instead of using a single line as a representation for the sequence, our main idea is to use a subspace parameterized by several orthonormal hyperplanes. We propose a novel learning-to-rank formulation to compute this subspace by minimizing an objective that jointly provides a low-rank approximation to the input data, while also preserves their temporal order in the subspace. The low-rank approximation helps capture the essential properties of the data that are useful for summarizing the action. Further, the temporal order is captured via a quadratic ranking function thereby capturing non-linear dependencies between the input data channels. Specifically, in our formulation, the temporal order is encoded as increasing lengths of the projections of the input data onto the subspace. 

While our formulation provides several advantages for temporal pooling, it leads to a difficult non-convex optimization problem, 
due to the orthogonality constraints. Fortunately, we show that the subspaces in our formulation satisfy certain mathematical properties, and thus can be cast as a problem on the so called~\emph{Grassmann manifold}, for which there exists efficient Riemannian optimization algorithms. We propose to use a conjugate gradient descent algorithm for our problem which is often seen to converge fast.
\done{\MH{I think when you say subspaces you have the linear combination property. I have seen papers saying frames to refer to an orthonormal matrix (a point on the Stiefel. I think saying subspaces in the abstract is OK, but here if you want to say something about why having orthogonality leads to Grassmannian, you need to stick to frames}}

We provide experiments on several popular action recognition datasets, preprocessed by extracting features from the fully-connected layers of VGG-net~\cite{simonyan2014very}. Following the standard practice, we use a two stream network~\cite{simonyan2014two} trained on single RGB frames and 20-channel optical flow images. Our experimental results show that the proposed scheme is significantly better at capturing the temporal structure of CNN features in action sequences compared to conventional pooling schemes or the basic form of rank-pooling~\cite{fernando2017rank}, while 
also achieving state-of-the-art performances. 

Before moving on, we summarize the main contributions of our work:
\begin{itemize}
	\item We propose a novel learning-to-rank formulation for capturing the temporal evolution of actions in video sequences by learning subspaces.
	\item We propose an efficient Riemannian optimization algorithm for solving our objective.
	\item We show that subspace representation on CNN features is highly beneficial for action recognition.
	\item We provide experiments on standard benchmarks demonstrating state-of-the-art performance.
\end{itemize}

\section{Related Work}
\label{sec:related_work}
Training of convolutional neural networks directly on long video sequences is often computationally prohibitive. Thus, various simplifications have been explored to make the problem amenable, such as using 3D spatio-temporal convolutions~\cite{tran2015learning}, recurrent models such as LSTMs or RNNs~\cite{donahue2014long,du2015hierarchical}, decoupling spatial and temporal action components via a two-stream model~\cite{simonyan2014two,feichtenhofer2016convolutional}, early or late fusion of predictions from a set of frames ~\cite{karpathy2014large}. While 3D convolutions and recurrent models can potentially learn the dynamics of actions in long sequences, training them is difficult due to the need for very large datasets and volumetric nature of the search space in a structurally-complex domain. Thus, in this paper, we focus on late fusion techniques on the CNN features generated by a two-stream model, and refer to recent surveys for a review of alternative schemes~\cite{Herath20174}.

Typically, the independent action predictions by a CNN along the video sequence is averaged or fused via a linear SVM~\cite{simonyan2014two} without considering the temporal evolution of the CNN features. Rank-pooling~\cite{fernando2015modeling} demonstrates better performances by accounting for the temporal information. They cast the problem in a learning-to-rank framework and propose an efficient algorithm for solving it via support-vector regression. While, this scheme uses hand-crafted features, extensions are explored in Fernando et al.~\cite{fernando2016discriminative,Fernando:ICML2016,su2016hierarchical} in a CNN setting via end-to-end learning. However, training such a deep architecture is slow as it requires computing the gradients of a bi-level optimization loss~\cite{gould2016differentiating}. This difficulty can be circumvented via early-fusion of the frames as described in Bilen et al.~\cite{bilen2016dynamic}, Wang et al.~\cite{dynamic_flow} by pooling the input frames or optical flow images, however one needs to solve a very high-dimensional ranking problem (with dimensionality equal to the size of input images), which may be slow. Instead, in this paper, we propose a generalization of the original ranking formulation~\cite{fernando2015modeling} using subspace representations and show that our formulation leads to significantly better representation of the dynamic evolution of actions, while being computationally cheap.

There have been approaches using subspaces for action representations in the past. These methods are developed mostly for hand-crafted features and thus their performances on CNN features are not thoroughly understood. For example, in the method of Li et al~\cite{Li2011}, it is assumed that trajectories from actions evolve in the same subspace, and thus computing the subspace angles may capture the similarity between activities. In contrast, we learn subspaces over more general CNN features and constrain them to capture dynamics. In Le et al.~\cite{Le2011}, the standard independent subspace analysis algorithm is extended to learn invariant spatio-temporal features from unlabeled video data. Principal components analysis and its variants have been suggested for action recognition in Karthikeyan et al.~\cite{Karthikeyan2011} using multi-set partial least squares to capture the temporal dynamics. This method also uses probabilistic subspace similarity learning proposed by~ Moghaddam et al.~\cite{Moghaddam1997} to learn intra-action and inter-action models. An adaptive locality preserving projection method is proposed in Tseng et al.~\cite{Tseng2012} to obtain a low-dimensional spatial subspace in which the linear structure of the data (such as that arising out of human body shape) is preserved. 

Similar to our proposed approach, O'Hara et al.~\cite{OHara2012} introduce a subspace forest representation for action recognition that considers each video segment as points on a Grassmann manifold and a random-forest based approximate nearest neighbor scheme is used to find similar videos. Subspaces-of-Features, formed from local spatio-temporal features, is presented in Raytchev et al.~\cite{Raytchev2013}, and uses Grassmann SVM kernels~\cite{Shawe-Taylor2004} for classification. A framework using multiple orthogonal classifiers for domain adaptation is presented in Etai and Wolf~\cite{littwin2016multiverse}. Similar kernel based recognition schemes are also proposed in Harandi et al.~\cite{harandi2013kernel} and Turaga et al.~\cite{turaga2011statistical}. In contrast, we are the first to propose subspace representations on CNN features for action recognition in a joint framework that includes non-linear chronological ordering constraints to capture the temporal evolution of actions.

\section{Proposed Method}
\label{sec:proposed_method}
Let $X=\seq{x_1, x_2, \cdots, x_n}$ be a sequence of $n$ consecutive data features, each $x_t\in\reals{d}$, produced by some dynamic process at discrete time instances $t$. In case of action recognition in video sequences using a two-stream CNN model, $X$ represents a sequence of features where each $x_t$ is the output of some CNN layer (for example, fully-connected FC6 of a VGG-net as used in our experiments) from a single RGB video frame or a small stack of consecutive optical flow images (similar to~\cite{simonyan2014two}). 

Our goal in this paper is to generate a compact representation for $X$ that summarizes the human action category and could be used for recognition of human actions from video. Towards this end, we assume that the per-frame features $x_t$ encapsulates the action properties of a frame (such as local dynamics or object appearance), and such features across the sequence captures the dynamics of the action as characterized by the temporal variations in $x_t$. That is, we assume the features are generated by a function $g$ parameterized by time $t$:
\begin{equation}
x_t = g(t),
\end{equation}  
where $g$ abstracts the action dynamics and produces the action feature for every time instance. However, in the case of real-world actions in arbitrary video sequences, finding such a generator $g$ is not viable. Instead, using the unidirectional nature of time, we impose an order to the generated features as suggested in~\cite{fernando2015modeling,fernando2017rank}, where it is assumed that the projections of $x_t$ onto some line preserves the order. 

Given that the features $x_t$ are often high-dimensional (as the ones we use, which are from the intermediate layers of a CNN), it is highly likely that the information regarding actions inhabits a low-dimensional feature subspace (instead of a line). Thus, we could write such a temporal order as:
\begin{equation}
\enorm{U^Tx_{t}}^2  \leq \enorm{U^Tx_{t+1}}^2 - \eta,
\label{eq:order_constraints} 
\end{equation}   
where $U\in\stiefel(p, d)$ denotes the parameters of a $p$-dimensional subspace, usually called a ~\emph{frame} ($p\ll d$) and $\eta$ is a positive constant controlling the degree to which the temporal order is enforced. Such frames have orthonormal columns and belongs to the Stiefel manifold $\stiefel(p, d)$~\cite{edelman1998geometry}. Our main idea is to use $U$ to represent the sequence $X$. To this end, we propose the following formulation for obtaining the low-rank subspace $U$ from $X$ given a rank $p$ as follows:
\begin{align}
\label{eq:low-rank}
\min_{U\in\stiefel(p,d)} L(U) \triangleq &\half\sum_{i=1}^n \enorm{x_i - UU^Tx_i}^2\\\nonumber
\subjectto\quad&  \enorm{U^T x_i}^2 \leq \enorm{U^T x_j}^2 - \eta,\quad \forall i<j.
\end{align}
In the above formulation, the objective seeks a $p$-rank approximation to $X$. Note that the Stiefel manifold enforces the property that $U$ has orthogonal columns, \ie, $U^TU=\eye{p}$, the $p\times p$ identity matrix.

\subsection{Properties of Formulation}
\label{sec:properties}
In this subsection, we explore some properties of this formulation that allows for its efficient optimization.
\noindent\paragraph{Invariance to Right-Action by Orthogonal Group:}
Note that our formulation in~\eqref{eq:low-rank} can be written as $L(U) = H(UU^T)$ for some function $H$. This means that for any matrix $R$ in the orthogonal group $\ortho(p)$, $L(UR) = H(URR^TU) = H(UU^T)$. This implies that all points of the form $UR$ are minimizers of $L(U)$. Such a set forms an equivalence class of all linear subspaces that can be generated from a given $U$ and is a point in the so called \emph{Grassmann} manifold $\grass(p,d)$. Thus, instead of minimizing over the Stiefel manifold, we could optimize the problem on the more general Grasssmann manifold.

\noindent\paragraph{Idempotence:} While, the objective in~\eqref{eq:low-rank} appears to be a quartic (fourth-order) function in $U$, it can be shown to be reduced to a quadratic objective as follows. Observe that the matrix $P=(I-UU^T)$ is symmetric idempotent, $\ie$, $P^TP=PP=P^2=P$. This implies that we can simplify the objective as follows:
\begin{equation}
\enorm{x_i - UU^Tx_i}^2 = \trace(x_ix_i^T(\eye{d}-UU^T)).
\label{eq:idempotent}
\end{equation} 
Unfortunately, the objective is concave and the overall formulation remains non-convex due to the orthogonality constraints on the subspace. 

Using the above simplifications, introducing slack variables, and rewriting the constraints as hinge-loss, we can reformulate~\eqref{eq:low-rank} and present our~\emph{generalized rank pooling} (GRP) objective as follows:
\begin{align}
&\min_{\substack{U: U\in\grass(p,d)\\ \xi\geq 0}}\hspace*{-0.3cm} F(U) \triangleq \half\sum_{i=1}^n \trace\left(x_ix_i^T(\eye{d}-UU^T)\right)\! +\! C\sum_{i<j}\xi_{ij}\nonumber\\
&\ \ +\frac{\lambda}{2}\sum_{i<j}\max(0, \enorm{U^T x_i}^2 - \enorm{U^T x_j}^2 + \eta - \xi_{ij}),
\end{align}
where $\lambda\rightarrow \infty$ is a regularization parameter and $\xi$ are non-negative slack variables.

\subsection{Efficient Optimization}
The optimization problem $F(U)$ can be solved via Riemannian conjugate gradient on the Grassmann manifold. The gradient of the objective at the $k$-th conjugate gradient step has the following form:
\begin{equation}
\nabla_U F(U_k) = \left[\sum_{\substack{\left(\forall i<j\right)\wedge\\ \enorm{U_k^Tx_j}^2-\enorm{U_k^Tx_i}^2\leq \eta-\xi_{ij}}}\hspace*{-1.5cm}\lambda\left(x_ix_i^T -x_jx_j^T\right) - XX^T\right] U_k,
\end{equation}
where the summation is over all constraint violations at a given iteration. Note that the complexity of this gradient computation is $\bigoh(d^2)$ where $d$ is the dimensionality of $x_i$, which may be expensive. Instead, below we propose a cheaper expression that leads to the same gradient.  

Suppose $V\in\set{0,1}^{n\times n}$ be a binary upper-triangular matrix whose $ij$-th $(j>i)$ entry describes if the points $x_i$ and $x_j$ violate the ordering constraints given $U_k$. Then, we can rewrite the above gradient as:
\begin{align}
\nabla_U F(U_k) & = \left[\sum_{i}\eta_{i} x_i(x_i^TU_k)\right] - X(X^T U_k),\\ \text{where } &\nu_i = \left(\sum V_{(i,:)}  - \sum V_{(:,i)}\right),
\end{align}
where $V_{(i,:)}$ and $V_{(:,j)}$ stand for the $i$-th row and $j$-th column of $V$, respectively. The complexity of computing $\nu_i$ is $\bigoh(n)$, and the cost of computing the gradient is reduced to $\bigoh(n + np)$.

\subsection{Incremental Subspace Estimation}
\label{sec:convex_reformulation}
The formulation introduced in~\eqref{eq:low-rank} estimates all the subspaces together, which may be expensive when working with high-dimensional subspaces. Instead, we show below that if we estimate the subspaces incrementally, that is one at a time, then each sub-problem can be solved more efficiently. To this end, suppose we have obtained the subspace $U_{q-1}\in\stiefel(p,q-1)$ and we are solving for the $q$-th basis vector $u_q$. The objective for finding $u_q$ can be recursively written as:
\begin{align}
&\quad\min_{\enorm{u_q}=1, \xi\geq 0}\quad\half\sum_{i=1}^n \enorm{\hat{x}_i - u_qu_q^T\hat{x}_i}^2 + C\sum_{i<j}  \xi_{ij}\nonumber\\
&\subjectto  \enorm{u_q^T\hat{x}_i}^2 \leq \enorm{u_q^T \hat{x}_j}^2  - \eta + \xi_{ij},\quad \forall i<j,\nonumber\\
&\hspace*{2cm}\hat{x}_i = x_i-U_{q-1}\langle U_{q-1}, x_i\rangle, \quad \forall i\nonumber\\
&\hspace*{3cm} u_qU_{q-1} = \mathbf{0}.
\label{eq:low_rank_inc}
\end{align}
In the above formulation, the main idea is to estimate each 1-dimensional subspace incrementally, and then subtract off the energy in $X$ associated with this subspace, thus generating $\hat{x}_i$ from $x_i$. Such unit subspaces are incrementally estimated (greedily) by following this procedure. As is clear, each solution of this objective solves a simpler problem under the quadratic objective, quadratic constraints, and a linear equality. Note that $U_{q-1}$ is a constant matrix when estimating $u_q$. However, the greedy strategy may lead to sub-optimal results, as is empirically also observed in Table~\ref{tab:all_comparisons-split1}. Albeit this greedy solution, the problem remains non-convex due to the concave objective and the difference-of-convex constraints.

\subsection{Conjugate Gradient on the Grassmannian}
\label{sec:opt_grassmannian}
As described in the last section, we cast the generalized rank pooling objective as an optimization problem with an orthogonality constraint, which can generally be written as
\begin{align}
	\label{eqn:opt_main_equ}
	\underset{U}{{\minimize}} &\quad F(U) \notag \\
	\subjectto\quad & U^TU = \eye{p},
\end{align}
where $F(U)$ is the desired cost function and $U\in\reals{d\times p}$. In the Euclidean space, problems of the form of~\eqref{eqn:opt_main_equ} are typically cast as eigenvalue problems. However, the complexity of 
our cost function prohibits us from doing so. Instead, we propose to make use of manifold-based optimization techniques.

Recent advances in optimization methods formulate problems with unitary constraints as optimization problems on
Stiefel or Grassmann manifolds~\cite{Edelman_1998,Absil_2008}. More specifically, the geometrically correct setting for the  
minimization problem in~\eqref{eqn:opt_main_equ} is, in general, on a Stiefel manifold. 
However, if the cost function $F(U)$ is independent from the choice of basis spanned by $U$, then the problem is on a Grassmann manifold. This is indeed what we showed in Section~\ref{sec:properties}.
We can therefore make use of Grassmannian optimization techniques, and, in particular, 
of Newton-type optimization, which we briefly review below.

Newton-type optimization, such as conjugate gradient (CG), over a Grassmannian is an iterative optimization routine that relies on 
the notion of Riemannian gradient. 
On $\GRASS{d}{p}$, the gradient is expressed as
\begin{equation}
\mathrm{grad}_U F(U) = (\mathbf{I}_d - UU^T)\nabla_U(F),
\label{eqn:grad_euc_2_grass}
\end{equation}
where $\nabla_U(F)$ is the $d \times p$ matrix of partial derivatives of ${F}(U)$ 
with respect to the elements of $U$. This is computed in Eq.\eqref{eq:low_rank_inc} for our method.
The descent direction expressed by $\mathrm{grad}_UF(U)$ identifies a curve $\gamma(t)$ on the manifold, moving along it ensures a decrease in the cost function (at least locally).  
Points on $\gamma(t)$ are obtained by the exponential map. In practice, the exponential map is approximated locally by a retraction (see Chapter 4 in~\cite{Absil_2008} for definitions and detailed explanations).
In the case of the Grassmannian, this can be understood as forcing the orthogonality constraint while making sure that the cost function decreases. 

In our experiments, we make use of a conjugate gradient (CG) method on the Grassmannian. CG methods compute the new descent direction by combining the gradient at the current and the previous solutions. To this end, it requires transporting the previous gradient to the current point on the manifold which is achieved by the concept of Riemannian connections. On the Grassmann manifold, operations required for a CG method, have efficient numerical forms, which makes them well-suited to perform optimization on the manifold.

\subsection{Classification on the Grassmanian}
Once we obtain the subspace representation solving the GRP objective using manifold CG method, the next step is to train a classifier on these subspaces for action recognition. Since these subspaces are elements of the Grassmannian, we must use an SVM kernel defined on this manifold. To this end, there are several potential kernels~\cite{harandi2014expanding}, of which we use the exponential Projection metric kernel due to its empirical benefits on our problem as validated in Table~\ref{tab:kernel_comparisons}. For two subspaces $U_1$ and $U_2$, the exponential projection metric kernel $K$ has the following form:
\begin{equation}
K(U_1, U_2) = \exp\left(\beta \fnorm{U_1^TU_2}^2\right), \text{for } \beta>0.
\end{equation}

\section{Experiments}
\label{sec:expts}
This section evaluates the proposed ranking method on four standard benchmark datasets on activity recognition, namely (i) the JHMDB dataset~\cite{jhuang2013towards}, (ii) the MPII Cooking activities dataset~\cite{rohrbach2012database}, (iii) the HMDB-51 dataset~\cite{kuehne2011hmdb}, and the UCF101 dataset~\cite{soomro2012ucf101}. In all our experiments, we use the standard 16-layer Imagenet pre-trained VGG-net deep learning network~\cite{simonyan2014very}, which is then fine-tuned on the respective dataset and input modality, such as single RGB or a stack of 10 consecutive optical flow images. We provide the details and evaluation protocols for each of these datasets below. 

\paragraph{HMDB Dataset:} consists of 6766 videos from 51 different action categories. The videos are generally of low quality, with strong camera motion, and non-centered people. 

\paragraph{JHMDB Dataset:} is a subset of HMDB dataset consisting of 968 clips and 21 different action classes. The dataset was mainly created for evaluating the impact of human pose estimation for action recognition, and thus all videos contain humans whose body-parts are clearly visible. 

\paragraph{MPII Cooking Activities Dataset:} consists of high-resolution videos of activities in a kitchen related to cooking several dishes. In comparison to the other two datasets, the videos are captured by a static camera. However, the activities could be very subtle such as \emph{slicing} or \emph{cutting vegetables}, \emph{washing} or \emph{wiping plates}, etc. that needs to be recognized. There are 5609 video clips and 65 annotated actions.  

\paragraph{UCF101 Dataset:} contains 13320 videos distributed in 101 action categories. This dataset is different from the above ones in that it contains mostly coarse sports activities with strong camera motion and low resolution videos. 

\subsection{Evaluation}
The HMDB, UCF101, and JHMDB datasets use mean average accuracy over 3-splits as their evaluation criteria. The MPII dataset uses 7-fold cross-validation and reports results on mean average precision (mAP). For the latter, we use the evaluation code published with the dataset.

\subsection{Preprocessing}  
The JHDMB, HMDB, and UCF101 datasets are relatively low resolution and thus we resize the images to input sizes that are required by the standard VGG-net model (that is, 224x224). We use the TVL1 optical flow implementation in Opencv to generate the 10-channel stack of flow images, where each flow image is rescaled in the range 0-255, and then saved as a JPEG image, which is the standard practice. 

For the MPII dataset, as the videos are originally in very high-resolution, we use a set of morphological operations to crop regions of interest before resizing them to the CNN input size. To be specific, we first resize the images into half their resolution, followed by computing the absolute difference between the frames, and summing up the differences across the sequence. Next, we apply median filtering, dilation, and connected component analysis to generate binary activity masks, and crop the sequences to the smallest rectangle that includes all the valid components. Once the sequences are cropped to these regions of interest, we use them as inputs to the CNNs, and also use them to compute stacked flow images.

\subsection{Training CNNs}
As alluded to earlier, we use the two-stream model of~\cite{simonyan2014two}, but uses the VGG-net architecture as it has demonstrated significant benefits~\cite{He2015,feichtenhofer2016convolutional}. However, our methods are not restricted to any specific architecture and could be used for deeper models such as the ResNet~\cite{feichtenhofer2016spatiotemporal}. The two network streams are trained independently against the softmax cross-entropy loss. The RGB stream is fine-tuned from the ImageNet model, while the flow stream is fine-tuned from the UCF101 model publicly available as part of~\cite{feichtenhofer2016convolutional}. We used a fixed learning rate of $10^{-4}$ and an input batch size of 50 frames. The CNN training was stopped as soon as the loss on the validation set started increasing, which happened in about 6K iterations for the RGB stream and 40K iterations for the optical flow. For the HMDB and JHMDB dataset, we use the 95\% of the respective training set in each split for fine-tuning the models, and rest is used as the validation set. The MPII cooking activities dataset comes with a training, validation, and a test set. For the UCF101 dataset, we used the models from~\cite{feichtenhofer2016convolutional} directly.
  
\subsection{Results}
This section provides a systematic evaluation of the influence of various hyper-parameters in our model, namely (i) influence of the number of subspaces used in our model, (ii) influence of the threshold used in enforcing the temporal order, (iii) comparison of the performance difference FC6 and FC7 CNN layer outputs in GRP, and (iv) an analysis of various Grassmannian kernels. We use the JHMDB and MPII datasets as common test beds for this analysis. In the following, we use the notations FLOW to denote a 10-channel stack of flow images, and RGB to denote the single RGB images.

We use the rectified output of the fully-connected layer fc6 of the VGG-net for all our evaluations which are 4096 dimensional vectors. All the features are unit-normalized before applying the pooling. We use the MANOPT~\cite{boumal2014manopt} software package for implementing the Grassmannian conjugate gradient. We run 100 iterations of this algorithm in all our experiments. Unless otherwise specified, we use the projection metric kernel~\cite{hamm2008grassmann} for classifying the subspaces on the Grassmannian. As for FLOW + RGB, which combines the GRP results from both FLOW and RGB streams, we use sum of two separate projection metric kernels one from each modality for classification. 

\subsubsection{Number of Subspaces and Ranking Threshold}
\begin{figure}
	\centering
	\subfigure[]{\label{fig:inc_subspaces}\includegraphics[width=4cm,,trim={0cm 7.cm 1.5cm 7cm},clip]{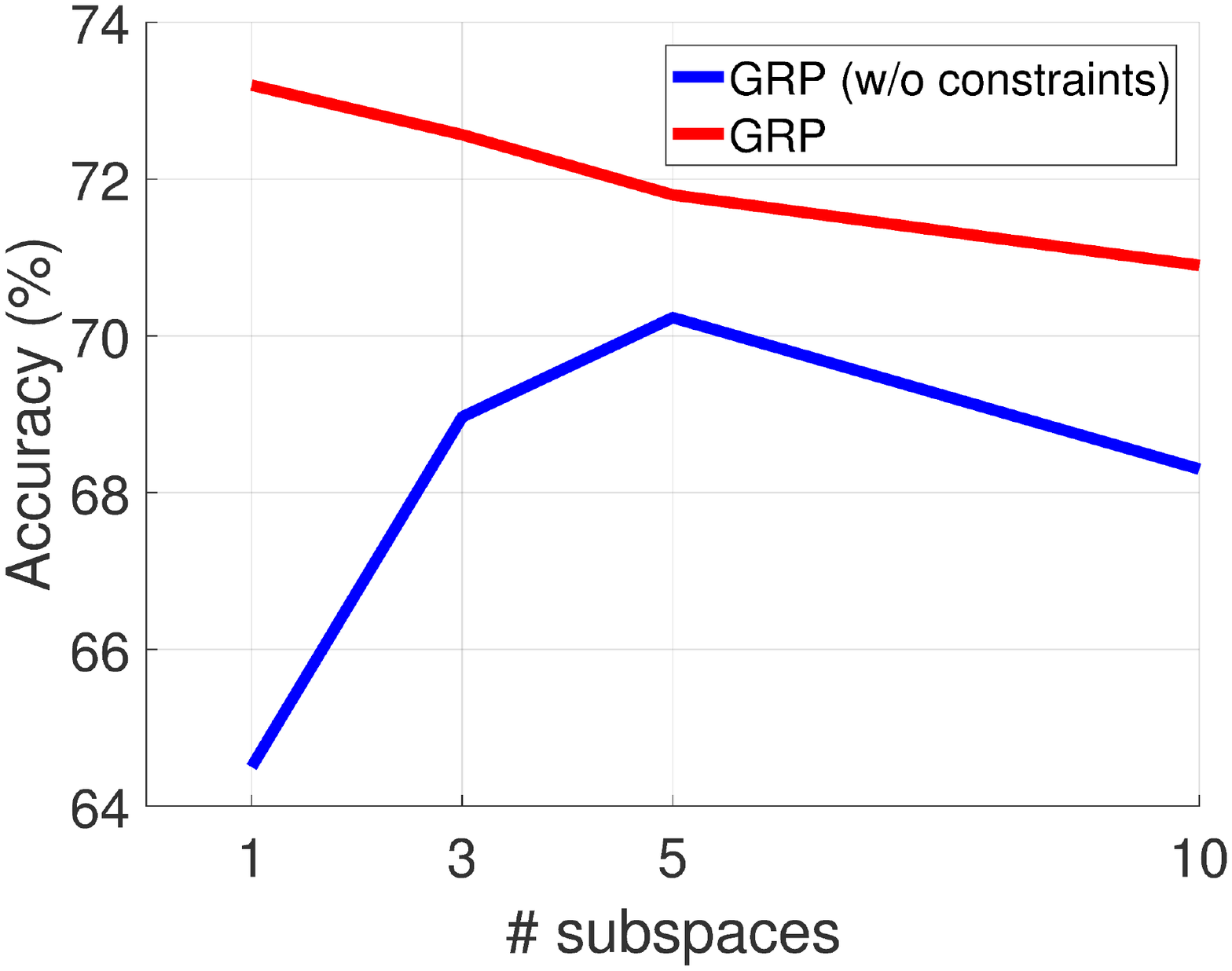}}
	\subfigure[]{\label{fig:ranking_threshold}\includegraphics[width=4cm,trim={0cm 7.cm 1.5cm 7cm},clip]{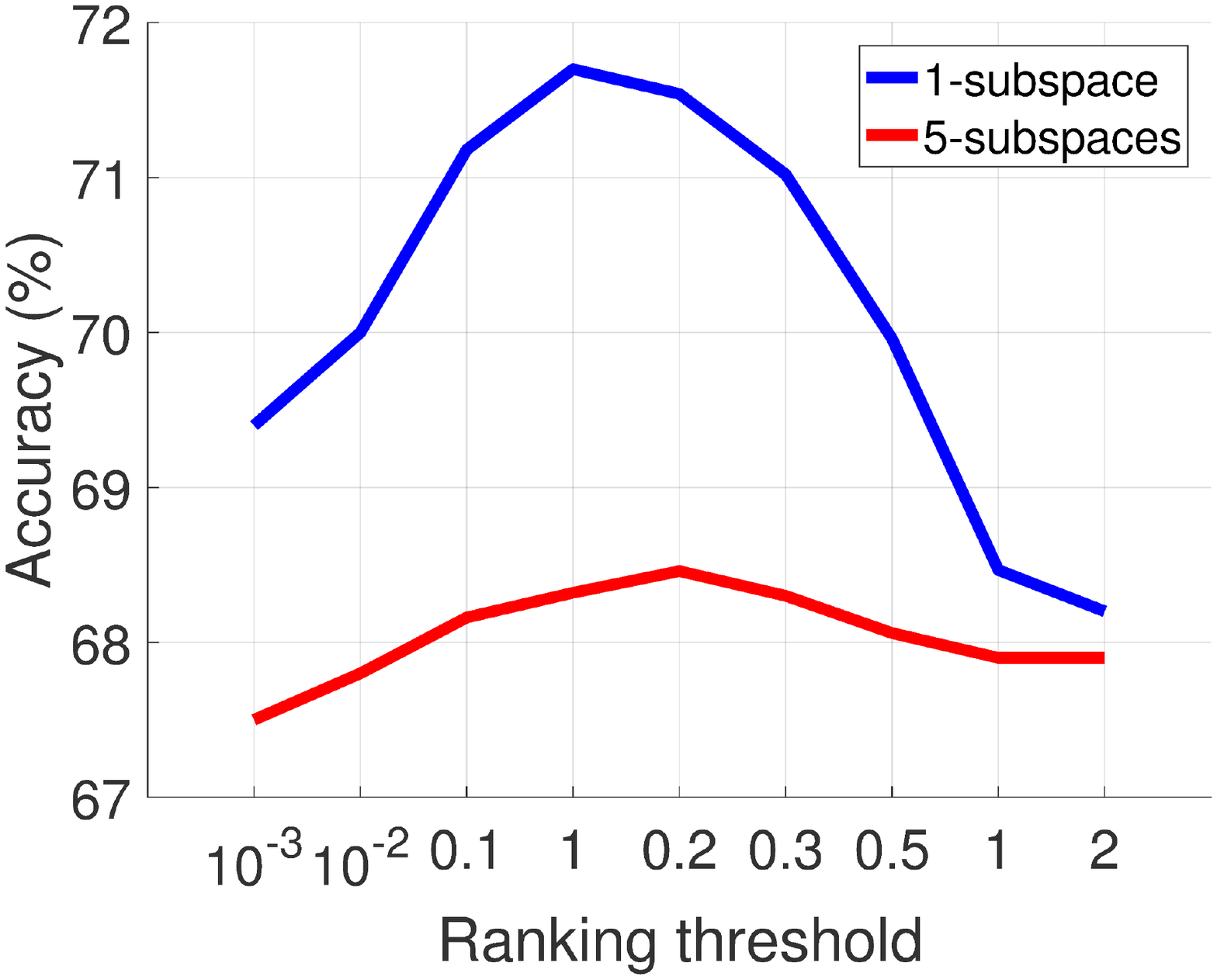}}
	\label{fig:inc_subspace_sizes}
	\caption{Left: Evaluation of accuracy against increasing subspace dimensionality (keeping fixed ranking threshold at 0.1). Right: Evaluation of accuracy against increasing ranking threshold $\eta$ keeping subspace dimensions fixed. Both results are on split 1 of JHMDB dataset and uses FLOW + RGB and does not use slack variables in the optimization.}
\end{figure}

In Figure~\ref{fig:inc_subspaces}, we compare the accuracy against an increasing number of subspaces used in the GRP formulation on the split-1 of the JHMDB dataset. We also compare the performance when not using the ranking constraints in the formulation. As is clear, temporal order constraints is beneficial, leading to about 9\% improvement in accuracy when using 1-2 subspaces, and about 3-5\% when using a larger number. This difference suggests that larger number of subspaces, which more closely approximates the input data, may be perhaps be capturing the background non-dynamics related features, which are not useful for classification.

To further validate the usefulness of our ranking strategy, we fixed the number of subspaces and increased the ranking threshold $\eta$ in~\eqref{eq:low-rank} from $10^{-3}$ to 2 in steps. Our plot in Figure~\ref{fig:ranking_threshold} shows that the accuracy of recognition increases significantly when the temporal order is enforced in the subspace reconstruction. However, when the number of subspaces is larger, this constraint does not help much, as is observed in the previous experiment as well. These two plots clearly demonstrate the correctness of our scheme and the usefulness of the ranking constraints in subspace representation. In the sequel, we use 2 subspaces in all our experiments, as it was seen most often to provide good results on the validation datasets. 

In Table~\ref{tab:ranking_comparison}, we compare the influence of the ranking constraints on the FLOW and the RGB channels separately. We note that these constraints have a bigger influence on the FLOW stream than on the RGB stream, implying that the dynamics are mostly being captured in the FLOW, as is obvious, while the RGB stream CNN is perhaps learning mostly the background context. A similar observation is also made in~\cite{dynamic_flow}. Nevertheless, it is noteworthy that these constraints does improve the performance even on the RGB stream.

\begin{table}[htbp]
	\centering
	\begin{tabular}{c|c|c}
		Method/Dataset &  FLOW  & RGB \\
		\hline
		MPII	   &  \small{mAP (\%)}   & \small{mAP (\%)}  \\
		\hline
		GRP (w/o constraints)     &  51 &  48.9   \\
		GRP-Grassmann             &  52.1  &  50.3   \\
		\hline
		JHMDB &\small{Avg.Acc.(\%)} & \small{Avg.Acc.(\%)}\\
		\hline
		GRP (w/o constraints) & 59.4 & 41.8\\
		GRP-Grassmann  & 64.2 & 42.5\\	
	\end{tabular}
	\caption{Comparison between the influence of GRP on FLOW and RGB separately on JHMDB and MPII datasets. These experiments use the split-1 of the respective datasets. Results of FLOW+RGB are in Table~\ref{tab:all_comparisons-split1}.}
	\label{tab:ranking_comparison}
\end{table}

\subsubsection{Choice of Grassmannian Kernel}
Another choice in our setup is the Grassmannian kernel to use. In Harandi et al.~\cite{harandi2014expanding}, a list of several useful kernels on this manifold is presented, each one behaving differently with respect to the application. To this end, we decided to evaluate the performance of these kernels on the subspaces generated from GRP. In Table~\ref{tab:kernel_comparisons}, we compare these kernels on the split-1 of MPII and the JHMDB datasets. We use the polynomial and RBF variants of the standard Projection Metric and the Binet-Cauchy distances. As depicted in the table, the linear kernel and the Binet-Cauchy kernels did not seem to perform well, but both the projection metric kernels seems to showcase significant benefits. 

\begin{table}[htbp]
	\centering
	\begin{tabular}{c|c|c}
		\small{Method/Dataset} &  \small{MPII (mAP\%)} & \small{JHMDB(Avg. Acc\%)}\\
		\hline
		Linear  &   24.2  & 46.6\\
		Poly. Proj. Metric     &  50.4 & 65.3 \\
		RBF Proj. Metric   &  52.1 &  66.8\\
		Poly. Binet-Cauchy &  33.6 & 40.0 \\
		RBF Binet-Cauchy      &  33.5 & 38.0
	\end{tabular}
	\caption{Comparison between the choice of different kernels for classification on the Grassmannian. We use the CNN features from the FLOW stream alone for this evaluation, using 2 subspaces.}
	\label{tab:kernel_comparisons}
\end{table}

\subsection{Comparison between CNN Features}
Next, we evaluate the usefulness of CNN features from the FC6 and FC7 layers. In Table~\ref{tab:cnn_feature_comparisons}, we provide this comparison on the split-1 of the JHMDB dataset, separately for the FLOW, RGB, and the combined streams. We see that consistently, the GRP on the FC6 layer performs better, perhaps it encodes more temporal information than the layers upper in the hierarchy. While, this posits that perhaps even lower intermediate layer features such as from Pool5 might be better. However, the dimensionality of these features is significantly higher and makes the GRP optimization harder in its current form.  

\begin{table}[htbp]
	\centering
	\begin{tabular}{c|c|c|c}
		Features &  FLOW & RGB & FLOW + RGB\\
		\hline
		JHMDB & Avg. Acc (\%) & Avg. Acc (\%)& Avg. Acc (\%)\\
		\hline
		FC6  &   64.2  & 42.5 & 73.8 \\
		FC7  &   63.4  & 40.3 & 72.0\\
		\hline
		MPII & mAP (\%) & mAP (\%)  & mAP (\%)\\
        \hline
		FC6  & 52.1 & 50.3 & 53.8\\
		FC7  & 45.6 & 46.5 & 50.7 
	\end{tabular}
	\caption{Accuracy comparison using FC6 and FC7 features.}
    \label{tab:cnn_feature_comparisons}
\end{table}
\subsection{Comparison between Pooling Techniques}
Now that we have a clear understanding of the behavior of GRP under disparate scenarios, we compare it against other popular pooling methods. To this end, we compare to (i) standard average pooling, (ii) Rank pooling ~\cite{fernando2015modeling}, which uses only a line for enforcing the temporal order, (iii) our GRP scheme but without ordering constraints, (iv) GRP-Grassmannian, which is our proposed scheme, and (v) our incremental reformulation of GRP, as described in Section~\ref{sec:convex_reformulation}. For Rank pooling, we use the publicly available code from the authors without any modifications. In Table~\ref{tab:all_comparisons-split1}, we provide these comparisons on the split-1 of all the four datasets. The results show that GRP is significantly better than average or Rank pooling consistently on all the four datasets. Further, surprisingly, we note that a low-rank reconstruction of the CNN features by itself provides a very good summarization of the actions useful for recognition. While, using subspaces for action recognition has been done several times in the past~\cite{harandi2013kernel,turaga2011statistical}, we are not aware of any work that shows these benefits on CNN features. However, using ranking constraints on low-rank subspaces leads to even better results. Specifically, there is about 7\% improvement on the JHMDB dataset, and about 4\% on the MPII dataset, 3\% on the HMDB datasets. We also note from these results that GRP-incremental works similar to GRP-Grassmannian, but shows slightly lower performance on an average. This is not surprising, given that it is a greedy method. 
 
\begin{table*}[htbp]
	\centering
	\begin{tabular}{c|c|c|c|c}
		Method/Dataset &  \small{MPII-mAP (\%)}  & \small{JHMDB-Avg.Acc.(\%)} & \small{HMDB-Avg. Acc.(\%)} & \small{UCF101-Avg. Acc. (\%)}\\
		\hline
		Avg. Pooling~\cite{simonyan2014two}          & 38.1  & 55.9  & 53.6  & 88.5\\
		Rank Pooling~\cite{fernando2015modeling}     & 47.2  & 55.2  & 51.4  & 63.8\\
		GRP (w/o constraints)                        & 50.1  & 67.5  & 62.2  & 90.4\\
		GRP-Grassman                                & 53.8  & 73.8  & 65.2  & 91.2\\
		GRP-incremental                              & 51.2  & 74.3  & 64.6  & 89.9\\
	\end{tabular}
	\caption{Comparison of various pooling techniques on the four datasets. We use the RGB+FLOW together for this evaluation on split-1.}
	\label{tab:all_comparisons-split1}
\end{table*}

\begin{figure}
	\centering    
	\includegraphics[width=9cm,trim={0cm 8cm 0cm 9cm},clip]{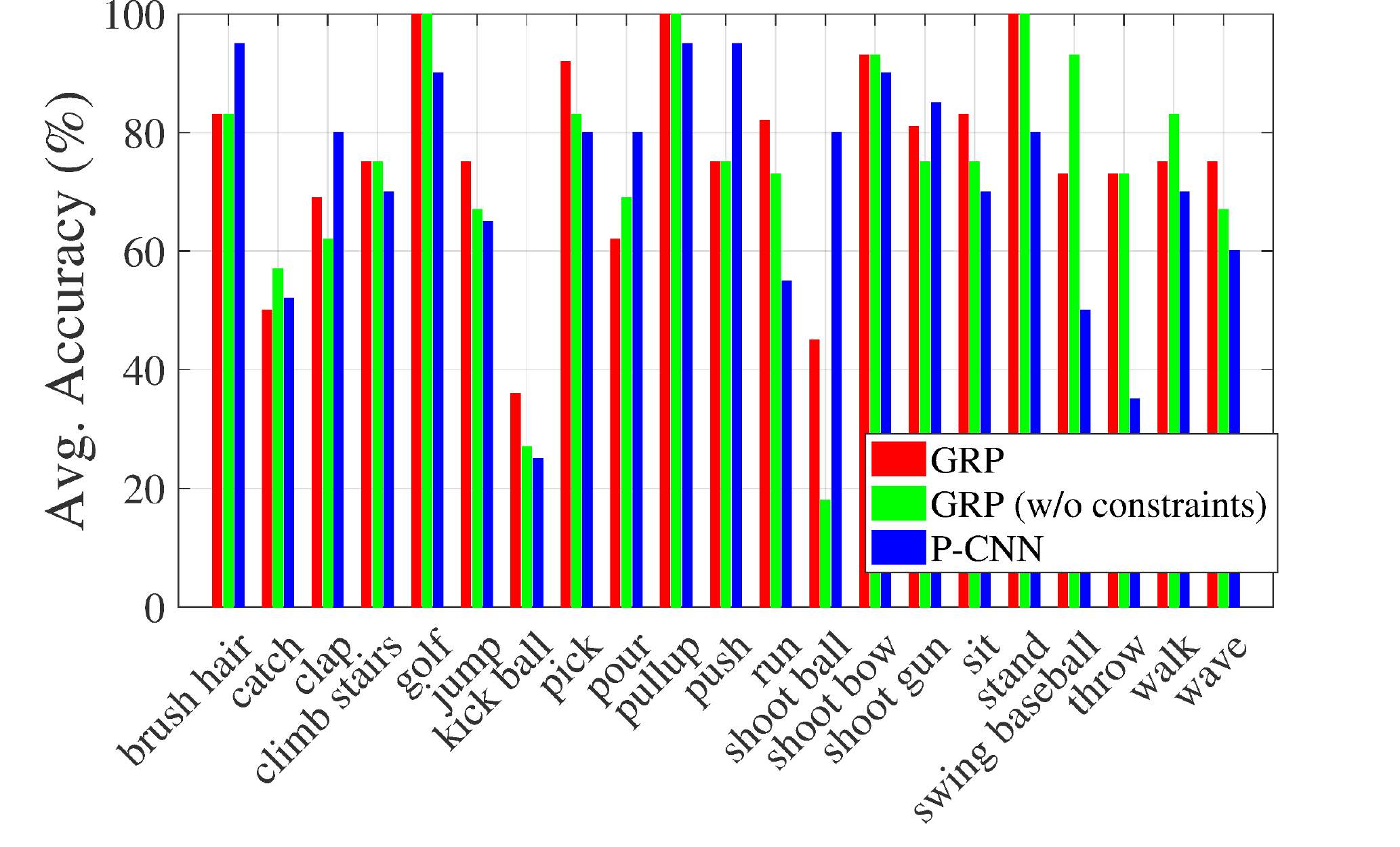}	
	\caption{Detailed comparison of the improvements afforded by GRP against the variant without ranking constraints and a recent state-of-the-art method~\cite{cheron2015p} on the JHMDB dataset (3-splits).}
    \label{fig:detailed_comparison_jhmdb}
\end{figure}

\subsection{Comparison to the State of the Art}
In Tables~\ref{tab:mpii_soa},~\ref{tab:jhmdb_soa},~\ref{tab:hmdb_soa}, and~\ref{tab:ucf_soa}, we compare GRP against state-of-the-art pooling and action recognition methods using CNNs and hand-crafted features. For all comparisons, we use the published results and follow the exact evaluation protocols. From the tables, it is clear that GRP outperforms the best methods on MPII and JHMDB datasets, while demonstrates promising results on HMDB and UCF101 datasets. For example, against rank pooling ~\cite{fernando2015modeling}, our scheme leads to significant benefits, by about 10-20\%  on MPII and JHMDB datasets (Table~\ref{tab:all_comparisons-split1}), while against dynamic images~\cite{bilen2016dynamic} without hand-crafted features it is better by 2-3\% on HMDB and UCF101 datasets. This shows that using subspaces leads to better characterization of the actions. Our results using VGG-16 model on these datasets are lower than the recent method in~\cite{feichtenhofer2016convolutional} that uses sophisticated residual deep models with intermediate stream fusion. Thus, we further analyzed the potential of our method using ResNet-152 features (fine-tuned from models shared as part of~\cite{feichtenhofer2016spatiotemporal}). Our results, using ResNet-152 models, on HMDB and UCF101 datasets in Tables~\ref{tab:hmdb_soa},~\ref{tab:ucf_soa} show that better CNN architectures do improve our model's performance significantly. We get about 7\% improvement over the VGG model and get state-of-the-art accuracy on HMDB-51, while showing competitive performance on UCF101. 

In Figure~\ref{fig:detailed_comparison_jhmdb}, we analyze the results of GRP, GRP-without constraints, and the recent P-CNN scheme~\cite{cheron2015p}. Out of 21 actions in this dataset, GRP outperforms P-CNN on 13. On 19 actions either GRP performs better or equal than the variant without constraints, thus substantiating its benefits.
\begin{table}[h]
	\centering    
	\begin{tabular}{c|c}
		Algorithm &  mAP(\%) \\
		\hline
		P-CNN + IDT-FV~\cite{cheron2015p} & 71.4 \\
		Interaction Part Mining~\cite{zhou2015interaction} & 72.4 \\
		Holistic + Pose~\cite{rohrbach2012database} & 57.9 \\
		Video Darwin~\cite{fernando2015modeling}    & 72.0 \\
		Semantic Features~\cite{zhou2014pipelining} & 70.5 \\
		Hierarchical Mid-Level Actions~\cite{su2016hierarchical} & 66.8\\
        Higher-order Pooling~\cite{hok} & 73.1\\
		\hline
		GRP (w/o constraints)   & 66.1 \\
		GRP & 68.4\\
		GRP + IDT-FV & \textbf{75.5}\\
	\end{tabular}    
	\caption{MPII Cooking Activities (7 splits)}	
    \label{tab:mpii_soa}
\end{table}
\begin{table}[h]
	\centering
	\begin{tabular}{c|c}
		Algorithm &  Avg. Acc. (\%) \\
		\hline
		P-CNN~\cite{cheron2015p} & 61.1 \\
		P-CNN + IDT-FV~\cite{cheron2015p} & 72.2 \\
		Action Tubes~\cite{gkioxari2015finding} & 62.5\\
		Stacked Fisher Vectors~\cite{peng2014action} & 69.03\\
		IDT + FV~\cite{wang2013action} & 62.8 \\    
        Higher-order Pooling~\cite{hok} & 73.3\\
		\hline       
		GRP (w/o constraints)  & 64.1 \\
		GRP  & 70.6 \\
		GRP + IDT-FV & \textbf{73.7}\\
	\end{tabular}    
	\caption{JHMDB Dataset (3 splits)}	
    \label{tab:jhmdb_soa}
\end{table}
\begin{table}[h]
	\centering
	\begin{tabular}{c|c}
		Algorithm &  Avg. Acc. (\%) \\
		\hline
		Two stream~\cite{simonyan2014two} & 59.4\\
		Spatio-Temporal ResNet~\cite{feichtenhofer2016spatiotemporal} & 70.3\\			Temporal Segment Networks~\cite{Wang2016} & 69.4\\
		TDD + IDT-FV~\cite{wang2015action} & 65.9\\		
		Dynamic Image + IDT-FV~\cite{bilen2016dynamic} & 65.2\\
        Hier. Rank Pooling + IDT-FV~\cite{fernando2016discriminative} & 66.9\\        
        Dynamic Flow + IDT-FV~\cite{dynamic_flow} & 67.4\\
		\hline
		GRP (w/o constraints)  & 63.1\\
		GRP  & 65.4 \\
		GRP + IDT-FV & 67.0\\
        GRP + IDT-FV (ResNet-152) & \textbf{72.1}
	\end{tabular}    
	\caption{HMDB Dataset (3 splits)}	
    \label{tab:hmdb_soa}    
\end{table}

\begin{table}[h]
	\centering    
	\begin{tabular}{c|c}
		Algorithm &  Avg. Acc. (\%) \\
		\hline
		Two stream~\cite{simonyan2014two} & 88.0\\
		Spatio-Temporal ResNet~\cite{feichtenhofer2016spatiotemporal} & \textbf{94.6}\\
        Temporal Segment Networks~\cite{Wang2016} & 94.2\\
		TDD + IDT-FV~\cite{wang2015action} & 91.5\\
		C3D + IDT-FV~\cite{tran2015learning} & 90.4\\		
        Dynamic Image + IDT-FV~\cite{bilen2016dynamic} & 89.1\\
        Hier. Rank Pooling + IDT-FV~\cite{fernando2016discriminative} & 91.4\\     
        Dynamic Flow + IDT-FV~\cite{dynamic_flow} & 91.3\\        
		\hline		
		GRP(w/o constraints)   & 90.1\\
		GRP  & 91.9\\
		GRP + IDT-FV & 92.3 \\
        GRP + IDT-FV (ResNet-152) & 93.5
	\end{tabular}    
	\caption{UCF101 Dataset (3 splits)}	
    \label{tab:ucf_soa}    
\end{table}
\vspace*{-0.6cm}

\section{Conclusions}
\label{sec:conclude}
We presented a novel algorithm, generalized rank pooling, to summarize the action dynamics in video sequences. Our main proposition was to use the parameters of a low-rank subspace as the pooled representation, where the deep learned features from each frame of the sequence is assumed to preserve their temporal order in this subspace. As such subspaces belong to the Grassmannian, we proposed an efficient conjugate gradient optimization scheme for pooling. Extensive experiments on four action recognition datasets demonstrated the advantages of our scheme.

\\
\noindent\textbf{Acknowledgements:} {This research was supported by the Australian Research Council (ARC) through the Centre of
Excellence for Robotic Vision (CE140100016). AC thanks the National Computational Infrastructure (NCI) for the support in experiments.}

{\small
\bibliographystyle{ieee}
\bibliography{genrankpool_bib}
}

\end{document}